\documentclass[letterpaper, 10 pt, conference]{ieeeconf}  

\IEEEoverridecommandlockouts                              

\usepackage{amsmath}
\usepackage{amsfonts}
\usepackage{url}
\usepackage{booktabs} 	
\usepackage{xspace}
\usepackage{pifont}
\usepackage[abspath]{currfile}
\usepackage[mode=buildnew]{standalone}

\usepackage{graphicx}
\usepackage{multirow}
\usepackage{color, colortbl}
\usepackage{wrapfig}
\usepackage{arydshln}	

\usepackage{tikz}
\usepackage{textcomp}
\usepackage{hyperref}
\usepackage{lipsum}

\newcommand{\reftab}[1]{Tab.~\ref{tab:#1}}

\newcommand{\boldparagraph}[1]{\vspace{0.1em}\noindent{\bf #1}}

\definecolor{LightCyan}{rgb}{0.88,1,1}

\newcommand\copyrighttext{%
  \footnotesize \textcopyright 2023 IEEE. Personal use of this material is permitted.
  Permission from IEEE must be obtained for all other uses, in any current or future 
  media, including reprinting/republishing this material for advertising or promotional 
  purposes, creating new collective works, for resale or redistribution to servers or 
  lists, or reuse of any copyrighted component of this work in other works. 
  }
\newcommand\copyrightnotice{%
\begin{tikzpicture}[remember picture,overlay]
\node[anchor=south,yshift=10pt] at (current page.south) {\fbox{\parbox{\dimexpr\textwidth-\fboxsep-\fboxrule\relax}{\copyrighttext}}};
\end{tikzpicture}%
}

\newcommand\acceptancetext{%
  \footnotesize Accepted for publication in IEEE International Conference on Robotics and Automation (ICRA), 2023
  }
\newcommand\acceptancenotice{%
\begin{tikzpicture}[remember picture,overlay]
\node[anchor=north,yshift=-10pt] at (current page.north)
{\parbox{\dimexpr\textwidth-\fboxsep-\fboxrule\relax}
{\acceptancetext}};
\end{tikzpicture}%
}

\title{\LARGE \bf
Learning-based Relational Object Matching Across Views
}

\author{Cathrin Elich$^{1,2}$, Iro Armeni$^{3}$, Martin R. Oswald$^{4}$, Marc Pollefeys$^{3,5}$, Joerg Stueckler$^{1}$
\thanks{*This work has been supported by Cyber Valley and the Max
Planck Society. We are grateful to the Max Planck ETH Center for Learning Systems for supporting Cathrin Elich.}
\thanks{$^{1}$Cathrin Elich and Joerg Stueckler are with Embodied Vision Group, Max Planck Institute for Intelligent Systems, Tuebingen, Germany
        {\tt\small firstname.lastname@tue.mpg.de}}%
\thanks{$^{2}$Cathrin Elich is with the Max Planck ETH Center for Learning Systems}
\thanks{$^{3}$Iro Armeni and Marc Pollefeys are with the Computer Vision and Geometry Lab, ETH Zurich, Switzerland
{\tt\small armeni@ibi.baug.ethz.ch, marc.pollefeys@inf.ethz.ch}}
\thanks{$^{4}$Martin R. Oswald is with University of Amsterdam, Netherlands
        {\tt\small m.r.oswald@uva.nl}}
\thanks{$^{5}$Marc Pollefeys is with Microsoft Mixed Reality and AI Lab, Zurich, Switzerland
}
}

\begin{document}

\maketitle
\thispagestyle{empty}
\pagestyle{empty}
\acceptancenotice
\copyrightnotice
\vspace{-\baselineskip}

\begin{abstract}
Intelligent robots require object-level scene understanding to reason about possible tasks and interactions with the environment.
Moreover, many perception tasks such as scene reconstruction, image retrieval, or place recognition can benefit from reasoning on the level of objects.
While keypoint-based matching can yield strong results for finding correspondences for images with small to medium view point changes, for large view point changes, matching semantically on the object-level becomes advantageous.
In this paper, we propose a learning-based approach which combines local keypoints with novel object-level features for matching object detections between RGB images.
We train our object-level matching features based on appearance and inter-frame and cross-frame spatial relations between objects in an associative graph neural network.
We demonstrate our approach in a large variety of views on realistically rendered synthetic images.
Our approach compares favorably to previous state-of-the-art object-level matching approaches and achieves improved performance over a pure keypoint-based approach for large view-point changes.
\end{abstract}


\section{Introduction}
Object-centric representations are essential for robots that act in their environment, for instance, to perceive obstacles for navigation, or to plan object manipulation actions.
Several approaches have been proposed that reason about scenes on the object-level for static and dynamic scene reconstruction~\cite{salas-moreno2013_slampp,strecke2019_emfusion,hughes2022_hydra}.
When tracking multiple objects in image sequences, or recognizing if places are revisited for loop closure detection, the ability to match objects between views of the environment becomes important.
Especially for large baselines changes between camera viewpoints, classical keypoint matching approaches~\cite{schoenberger2016sfm} often break down due to drastic appearance changes and partial occlusions.

In this paper, we propose a novel approach for matching object detections between images by leveraging both spatial and appearance features of objects and spatial relations between the objects.
In our approach, we combine keypoints with object-level feature correspondences to use the best of both worlds to match objects across small to large view point changes. 
On the object-level, the matching features are determined by an associative graph neural network that reasons on spatial and appearance features of object detections in both input images.
To incorporate spatial knowledge into our matching pipeline, we additionally train our features on semantic and spatial auxiliary tasks such as inferring object class, 3D position, and relative distance between objects. 
By this, objects can be matched based on semantic and spatial information across large view point changes.
This is in contrast to pure 2D keypoint-based matching approaches~\cite{sarlin20_superglue} which rely on local texture similarities and do not exploit object semantics.
We evaluate our approach on a realistically rendered indoor dataset~\cite{roberts21_hypersim} and demonstrate improved accuracy over state-of-the-art object matching approaches.
For large view-point changes, our approach also compares favorably in object matching to a baseline that uses the state-of-the-art keypoint matching method in accuracy and recall.
We provide an ablation study to analyze the contributions of the individual design choices in our method.

In summary, we make the following contributions:
\textbf{(1)}~We present a novel learning approach for learning relational object features suited for matching detected objects between image pairs.
Our approach uses an attentional graph neural network based on appearance and spatial features extracted from the object bounding boxes. 
\textbf{(2)} We combine our object-level features with learning-based keypoint matching to achieve state-of-the-art object matching for small to large view point changes.

\begin{figure*}[tb]
	\centering
	\includegraphics[width=\linewidth]{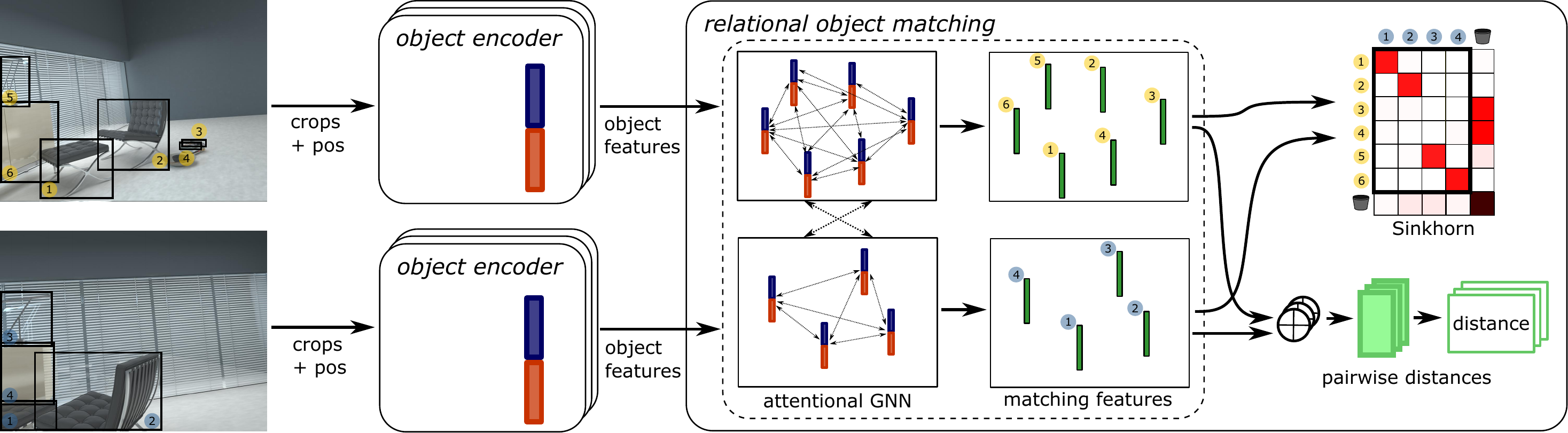}
	\caption{\textbf{Learning relational object matching (ROM) features.} Our approach learns object-level features for matching objects between images. An object encoder network extracts object-wise features which encode appearance and positional information from the object bounding boxes. The features are input to an Attentional Graph Neural Network (AGNN) with self attention (within image) and cross attention (across images). The AGNN yields matching features for each object from which pairwise matching scores for objects of both images are computed by taking the dot product. An approximation to the optimal partial assignment between objects is determined using the Sinkhorn algorithm. The set of objects is augmented with a dustbin to allow for unmatched objects due to occlusions and limited field of view. 
	Ground-truth object bounding boxes are used in the depicted example.
	}
	\label{fig:pipeline_object}
\end{figure*}

\section{Related Work}

Recently, several object-centric scene reconstruction approaches have been proposed that detect objects in images and aggregate the detections into scene representations like 3D object-level maps or scene graphs.
For instance, methods for incremental simultaneous localization and mapping have been proposed which detect, segment, and reconstruct static and moving objects~\cite{mccormac2018_fusionpp,ruenz2018_maskfusion,xu2019_midfusion,strecke2019_emfusion,ruenz2020_frodo}.
Scene graph-based approaches also estimate spatial and semantic relations between objects~\cite{armeni19_3dSceneGraph,wald20_3dSemSceneGraph,rosinol21_kimera,gothoskar21_3dp3}.
Common to approaches which create object-centric scene representations from images is that they require means to detect or segment objects in the images and to associate them between images or instances in the accumulated scene representation.
Typically this association is tackled using geometric data association.

In the multi-object tracking literature, data association is often performed using object-wise features extracted locally from the object bounding boxes which can aid in scenarios with occlusions~\cite{wojke2017_deepsort}. 
Other video tracking approaches associate objects implicitly in neural encoder-decoder or transformer architectures~\cite{liu2020_snippet,cong2021_transformer}.
For scene graph fusion from objects detected in RGB images, CSR~\cite{gadre2022_csr} learns encodings of object bounding boxes into feature vectors which are used to measure similarity of detected objects between views using cosine similarity. 
In Associative3D~\cite{qian2020_associative3d}, also feature embeddings per objects are learned which are used to match objects between views. 
Both \cite{schonberger2018_semantic} and \cite{speciale2018_consensus} leverage semantic and spatial information for finding 3D correspondences.
Recently, \cite{ma2022_virtual} propose to use human models to estimate correspondences among wide-baseline view changes.
Different to object associations, keypoint matching methods yield local image correspondences between pairs of images for moderate view point changes~\cite{sarlin20_superglue, sun2021_loftr, jiang2021_cotr}.
In our approach, inspired by SuperGlue~\cite{sarlin20_superglue}, we further process individual object embeddings in a graph neural network which relates objects within images and across images to encode image context.
Our approach can be an alternative or complementary approach for matching objects across views to geometric or keypoint-based approaches.

\section{Method}

\begin{figure*}[tb]
	\centering
	\includegraphics[width=0.8\linewidth]{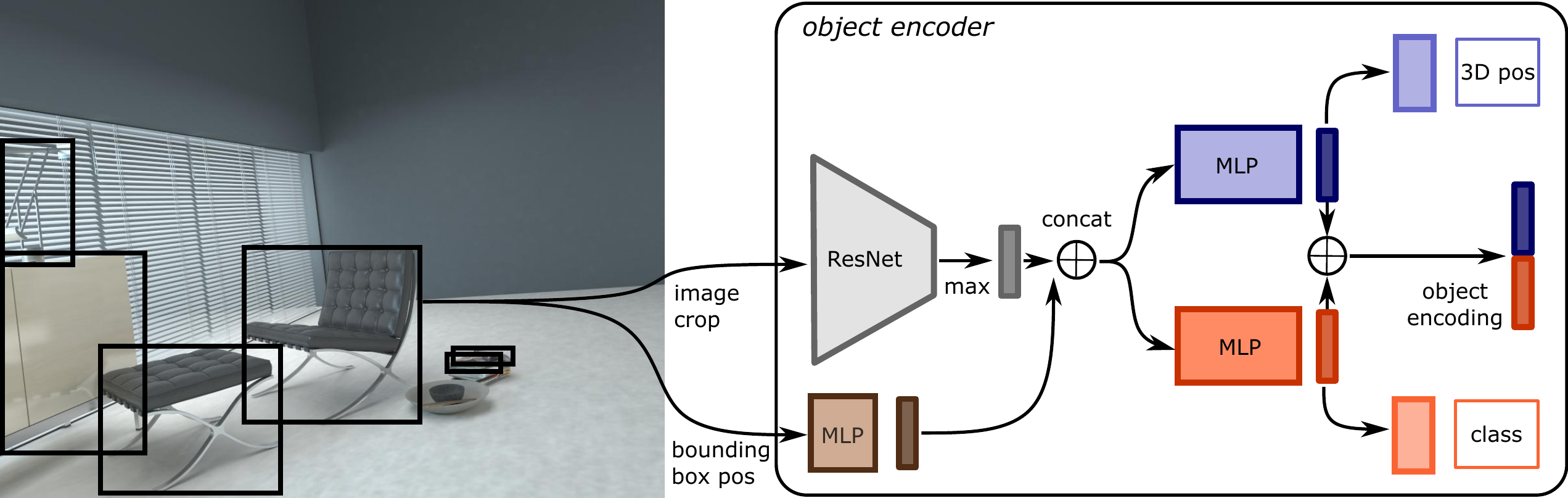}
	\caption{\textbf{Object encoder.} From object bounding boxes, a ResNet34 pretrained on ImageNet extracts features which are max-aggregated channelwise into appearance feature vectors per object. The appearance features are concatenated with an MLP-encoding of the bounding box coordinates of the object. This object input feature is further processed by two MLPs into view-independent and view-dependent features. The first is trained with a 3D object position side task, while the latter is trained for semantic object classification. Both features are concatenated into an intermediate object feature. }
	\label{fig:object_encoder}
\end{figure*}

Our approach finds matches of corresponding objects in a 3D scene.
Input to our method are pairs of images $I_1, I_2$ along with a set of object detections in each image.
The objects can be extracted with any object detector such as YOLO~\cite{redmon2016_yolo}, SSD~\cite{liu2016_ssd}, or Mask R-CNN~\cite{he2017_maskrcnn}.
Figs.~\ref{fig:pipeline_object} and~\ref{fig:object_encoder} illustrate our approach.
We extract deep encodings from each object detection.
The encodings are further processed in an attentional graph neural network which establishes context within each image as well as across images.
The resulting matching feature vectors are used to determine a similarity score for each possible pair of objects.
Differentiable Sinkhorn matching~\cite{knopp1967_sinkhorn,cuturi2013_sinkhorn} then finds a unique best mapping for each object to either objects or an outlier class (dustbin). 
The network is trained end-to-end using ground truth matches of objects on realistically rendered images of indoor scenes.
Finally, we propose an object matching pipeline which combines object-level features with learning-based keypoints (SuperGlue~\cite{sarlin20_superglue}) for matching in a wide range of view point changes.

\subsection{Object-Level Encoding}
In each image, we encode the objects specified by 2D bounding boxes into two kinds of feature vectors which capture view-dependent and view-independent object information.
We start from visual features~$\mathbf{f}_{i,\mathit{viz}}\in\mathbb{R}^{D_{\mathit{viz}}}$ for each object~$i$ which we determine by extracting ResNet34~\cite{he2016_resnet} features pretrained on ImageNet within the object bounding boxes.
The ResNet34 features are aggregated into a feature vector using a channel-wise max operation.
Additionally, an MLP~$g_{\mathit{loc}}$ calculates a feature vector $\mathbf{f}_{i,\mathit{loc}}\in\mathbb{R}^{D_\mathit{loc}}$ from the bounding box coordinates which we normalize by the image size.
We use ReLU activation functions in each hidden layer.
The visual and location features are concatenated into an input feature $\mathbf{f}_{i,\mathit{in}}:=(\mathbf{f}_{i,\mathit{viz}}, \mathbf{f}_{i,\mathit{loc}})$. 
We use 3D regression and semantic object classification side-tasks to train features which are expressive in view-dependent and view-independent object properties.
For the view-dependent features $\mathbf{f}_{i,\mathit{dep}} := g_{\mathit{dep}}(\mathbf{f}_{i,\mathit{in}})$, we apply an MLP on the input features and train a subsequent MLP~$g_{\mathit{pos}}$ to predict the 3D position parametrized by a 2D image offset $\Delta\mathbf{p}_i \in \mathbb{R}^2$ from bounding box center to image projection of the 3D position and distance $d_i \in \mathbb{R}$, similar to Total3D~\cite{nie2020_total3d}.
The view-independent features $\mathbf{f}_{i,\mathit{indep}}:=g_{\mathit{indep}}(\mathbf{f}_{i,\mathit{in}})$ are similarly extracted using an MLP, and another MLP $g_{\mathit{class}}$ extracts logits for semantic object classification.
We obtain the final object feature~$\mathbf{f}_{i}:=(\mathbf{f}_{i,\mathit{dep}}, \mathbf{f}_{i,\mathit{indep}}) \!\in\! \mathbb{R}^{D_\mathit{obj}}$ by concatenating the  view-dependent and view-independent features.

\subsection{Relational Object Matching}
In a subsequent processing step, we use an Attentional Graph Neural Network~(AGNN) as in~\cite{sarlin20_superglue} to obtain object encodings for matching.
Object encodings are nodes in the graph, while graph edges model relations between objects within each image and across the images.
The graph is fully connected and allows for interactions between all objects in both images.
We denote object encodings by $\mathbf{x}_{k,i}$ where~$k \in \{ 1,2 \}$ refers to the image of the object and $i$ specifies the object index. 
The object encodings 
$\mathbf{x}_{k,i,l} = \mathbf{x}_{k,i,l-1} + g_{\mathit{AGNN}}\big( ( \mathbf{x}_{k,i,l-1}, \mathbf{m}_{i.l} ) \big)$
 are refined in each AGNN stage~$l$ by passing message $\mathbf{m}_i$ aggregated from all other object nodes in the graph, i.e. within the same image and across images.
The encodings are initialized with the respective object features~$\mathbf{f}_i$.
The messages are determined by an attention mechanism.
An affine layer extracts values~$\mathbf{v}_j$ and keys $\mathbf{k}_j$ for each other object in the graph from the current object encodings $\mathbf{x}_{j}$.
Here, we drop the image index for simplicity of notation, since objects from both images are treated equally.
For the object to be updated, we also determine the query $\mathbf{q}_i$ using an affine layer for its current encoding.
The messages are computed by aggregating the values
$\mathbf{m}_i = \sum_{j:(i,j)\in E} \alpha_{ij} \mathbf{v}_j$
with attention weights~$\alpha_{ij} = \operatorname{softmax}_j\left( \mathbf{q_i}^\top \mathbf{k}_j \right)$ and graph edges~$E$.

The AGNN yields refined object features~$\mathbf{x}_{k,i} \in \mathbb{R}^{D_{\mathit{AGNN}}}$ for object~$i$ in image~$k$ which are used to compute a similarity score using a scalar product between the feature vectors of objects in different images.
In an additional 3D side task, we train the features to encode view-independent relative distance information between objects in each image.
To this end, the relative distance
$d_{k,ij} := g_{\mathit{dist}}\left((\mathbf{x}_{k,i},\mathbf{x}_{k,j})\right)$
is regressed from pairs of refined features in the same image~$k$ using an MLP.
For further details on the AGNN, please refer to~\cite{sarlin20_superglue}.
Finally, we use the Sinkhorn algorithm~\cite{knopp1967_sinkhorn,cuturi2013_sinkhorn} to compute a differentiable approximation to the optimal matching of objects based on their matching score.
Let $M,N$ be the number of objects in image 1 and 2, respectively.
Input to the Sinkhorn algorithm is the pairwise matching score of the objects determined by the scalar product of their feature vectors $\mathbf{S}_{ij}^{\mathit{obj}} = \mathbf{x}_{1,i}^\top \mathbf{x}_{2,j}$.
This leads to a $M \times N$ matrix~$\mathbf{S}^{\mathit{obj}}$.
As noted in~\cite{sarlin20_superglue}, using unnormalized feature vectors allows the network to learn prediction confidence implicitly.

Some objects might not be visible in both images due to occlusions or limited field-of-view.
To allow for objects being not visible and matchable between frames for the matching, we additionally include an outlier association by a dustbin class with a single learnable parameter for each object.
The augmented score matrix is denoted by $\overline{\mathbf{S}}^{\mathit{obj}} \in \mathbb{R}^{(M+1) \times (N+1)}$.
Finally, let $\mathbf{1}_{K} \in \mathbb{R}^K$ be a vector of ones in each dimension.
The optimal matching problem is to find unique matchings for each objects to either other objects or the dustbins for each image.
The dustbins are allowed to be associated multiple times.
This linear assignment problem is
\begin{multline}
    \underset{\overline{\mathbf{P}}\in \left[ 0, 1 \right]^{M+1 \times N+1}}{\operatorname{arg\,max}} \sum_{i,j} \overline{\mathbf{S}}^{\mathit{obj}}_{ij} \overline{\mathbf{P}}_{ij}\\
    \text{s.t. }\overline{\mathbf{P}} \mathbf{1}_{N+1} = \left( \mathbf{1}_M^\top N \right)^\top \text{ and } \overline{\mathbf{P}}^\top \mathbf{1}_{M+1} = \left( \mathbf{1}_N^\top M \right)^\top
\end{multline}
with assignment matrix $\overline{\mathbf{P}}$.
The Sinkhorn algorithm iteratively normalizes rows and columns of $\exp\left(\overline{\mathbf{S}}^{\mathit{obj}}\right)$ to arrive at $\overline{\mathbf{P}}$.
See~\cite{sarlin20_superglue} for additional details.

\subsection{Loss Function}

We train our model supervised using ground truth object bounding boxes and matches between pairs of images.
Our overall loss function 
\begin{equation}
    \mathcal{L} = \lambda_{\mathit{aff}} \mathcal{L}_{\mathit{aff}} + \lambda_{\mathit{cls}} \mathcal{L}_{\mathit{cls}} +
    \lambda_{\mathit{pos}} \mathcal{L}_{\mathit{pos}} +
    \lambda_{\mathit{rel}} \mathcal{L}_{\mathit{rel}} 
\end{equation}
is composed of four losses with corresponding weighting terms $\lambda_{\mathit{aff}}$, $\lambda_{\mathit{cls}}$, $\lambda_{\mathit{pos}}$ and $\lambda_{\mathit{rel}}$ accounting for affinity matching, classification, position, and relative distance.
We use the affinity loss~$\mathcal{L}_{\mathit{aff}}$ as in~\cite{sarlin20_superglue} which compares the predicted assignment matrix with the ground-truth assignment.
The loss $\mathcal{L}_{\mathit{cls}}$ is a per-class weighted category cross entropy loss for the classification predicted from the view-independent features.
The 3D position regression is trained with the loss
\begin{equation}
\begin{aligned}
\mathcal{L}_{\mathit{pos}} = 
  &\frac{1}{M} \Big( \sum_i (\mathbf{p}_{1,i} - \mathbf{p}_{1,i}^{gt})^2 + (d_{1,i} - d_{1,i}^{gt})^2 \Big) \\
  & + \frac{1}{N} \Big( \sum_j (\mathbf{p}_{2,j} - \mathbf{p}_{2,j}^{gt})^2 + (d_{2,i} - d_{2,i}^{gt})^2 \Big) 
\end{aligned}
\end{equation}
which measures the Euclidean distance between the predicted and the ground-truth position.
We do not consider objects for this loss if the offset from the center of the bounding box is large (more than the bounding box length in each dimension), since these objects are typically found at the image border and lack image content for reliable prediction. 
The error in the prediction of the relative distance between objects is measured using the loss 
\begin{equation}
    \mathcal{L}_{\mathit{rel}} = \sum_{k=1}^2 \sum_{i} \sum_{j, j \neq i} (d_{k,ij} - d_{k,ij}^{gt})^2.
\end{equation}

\subsection{Combined Object-Level and Keypoint-based Matching}

Classically, keypoints with local descriptors have been used to find correspondences between images.
While modern learning-based approaches such as SuperGlue~\cite{sarlin20_superglue} can achieve highly accurate results, finding correspondences across large view points still poses a challenge for these methods, since they do not take object-level information into account.
We thus propose to combine keypoint-based matching with our object-level feature matching.
To this end, we find keypoint matches between two images using SuperGlue as state-of-the-art learning-based keypoint detector. 
Each keypoint match is assigned to those object matches for which the keypoints are located within the bounding box of the objects.
The score~$\mathbf{S}_{ij}^{\mathit{kp}}$ of each object match for object~$i,j$ is determined by the logarithm of the count of keypoint matches between the objects.
An additional dustbin is added for each object with a score of~$1$.
The combined object matching score $\overline{\mathbf{S}} = \overline{\mathbf{S}}^{\mathit{obj}} + \alpha \, \overline{\mathbf{S}}^{\mathit{kp}}$ is a linear combination of the object-level and keypoint-based scores, from which we find the optimal assignment using the Sinkhorn algorithm.
We use~$\alpha = 100$ in our experiments which we chose based on the F1-score on the validation set.

\begin{table}[tb]
	\footnotesize
	\centering
	\caption{\textbf{Network Parameters.}}
	\begin{tabular}{llcc}
		\toprule
        MLP & hidden units, output dimensionality \\
        \midrule
        $g_{\mathit{loc}}$ & $[32, 64, 128]$ \\
        $g_\mathit{dep/ indep}$ & $[512, 256, 128, 128]$ \\
        $g_\mathit{pos}$ & $[256, 3]$\\
        $g_\mathit{class}$ & $[256, 40]$\\
        $g_\mathit{dist}$ & $[256, 1]$\\
        $g_\mathit{AGNN}$ & [(self)256, (cross)256, (self)256, (cross)256]\\
	 	\bottomrule
	\end{tabular}	
	\label{tab:params}
\end{table}
\begin{table}[tb]
	\scriptsize
	\centering
	\setlength{\tabcolsep}{1.3pt}
	\newcommand{\gcs}{\hspace{4pt}} 
	\caption{\textbf{Matching results for ground-truth detections as input. Top:} object-wise, \textbf{bottom:} frame-wise. Our approach performs best among trained object matching approaches, and outperforms keypoint-based matching (SuperGlue) in F1-score and recall for large viewpoint changes.}
	\label{tab:gtmatching}
	\begin{tabular}{lc@{\gcs}cccc@{\gcs}cccc@{\gcs}ccc}
		\toprule
         && \multicolumn{3}{c}{Easy} && \multicolumn{3}{c}{Hard} && \multicolumn{3}{c}{Very Hard}\\
		\cmidrule(l){2-5} \cmidrule(l){6-9} \cmidrule(l){10-13} 
	 	\bf object-wise && F1$\uparrow$ & Prec$\uparrow$ & Rec$\uparrow$ && F1$\uparrow$ & Prec$\uparrow$ & Rec$\uparrow$ && F1$\uparrow$ & Prec$\uparrow$ & Rec$\uparrow$ \\
 		\midrule
		SuperGlue~\cite{sarlin20_superglue}         && \bf 0.657 & \bf 0.811 & \underline{0.552} && 0.353& \bf 0.682 & 0.250 && 0.169 & \bf 0.574 & 0.099 \\[0.3pt]
		CSR~\cite{gadre2022_csr}                    && 0.357 & 0.404 & 0.320 && 0.209 & 0.235 & 0.188 && 0.189 & 0.191 & 0.187 \\
		Associative3D~\cite{qian2020_associative3d} && 0.259& 0.265 & 0.253 && 0.191 & 0.180 & 0.204 && 0.194 & 0.164 & 0.239\\ 
		\hdashline \noalign{\vskip 2pt}
		ROM features && 0.538 & 0.594 & 0.492  && \underline{0.358} & 0.431 & \underline{0.306} && \underline{0.308} & 0.381 & \underline{0.259}\\
		Ours && \underline{0.642} &  \underline{0.647} & \bf 0.638 && \bf 0.415 & \underline{0.450} & \bf 0.385 && \bf 0.332 & \underline{0.385} & \bf 0.292\\
	 	\midrule
	 	\midrule
	 	\bf frame-wise&& F1$\uparrow$ & Prec$\uparrow$ & Rec$\uparrow$  && F1$\uparrow$ & Prec$\uparrow$ & Rec$\uparrow$ && F1$\uparrow$ & Prec$\uparrow$ & Rec $\uparrow$ \\
		\midrule
		SuperGlue~\cite{sarlin20_superglue}          && \bf 0.740 & \bf 0.811 & \underline{0.589} && \underline{0.551} & \bf 0.671 & 0.278 && \underline{0.489} & \bf 0.557 & 0.099 \\[0.3pt] \hdashline \noalign{\vskip 2pt}
		CSR~\cite{gadre2022_csr}                     && 0.441 & 0.409 & 0.385 && 0.338 & 0.257 & 0.257 && 0.336 & 0.214 & 0.239\\
		Associative3D~\cite{qian2020_associative3d}  && 0.328 & 0.312 & 0.312 && 0.274 & 0.228 & 0.264 && 0.284 & 0.204 & 0.282\\
		 \hdashline \noalign{\vskip 2pt}
		ROM features && 0.625 & 0.628 & 0.553 && 0.493 & 0.473 & \underline{0.381} && 0.469 & 0.402 & \underline{0.326}\\
		Ours && \underline{0.723} & \underline{0.688} & \bf 0.694 && \bf 0.552 & \underline{0.500} & \bf 0.464 && \bf 0.498 & \underline{0.419} & \bf 0.367\\
	 	\bottomrule
	\end{tabular}	
\end{table}

\begin{table*}[t]
	\scriptsize
	\centering
	\setlength{\tabcolsep}{1.5pt}
	\newcommand{\gcs}{\hspace{10pt}} 
	\caption{\textbf{Average results on auxiliary tasks} using ground-truth object detections. 
	Our approach achieves better or on-par results than Associative3D for 3D subtasks, while performing worse for classification.}
	\label{tab:sidetasks}
	{\resizebox{1.\textwidth}{!}{
	\begin{tabular}{lc@{\gcs}cc@{\gcs}ccccc@{\gcs}cccc}
		\toprule
		&& \multicolumn{1}{c}{Classification} && \multicolumn{4}{c}{3D Position} && \multicolumn{4}{c}{3D Distance} \\
		\cmidrule(l){2-3} \cmidrule(l){4-8} \cmidrule(l){9-13}
	 	&&  Acc $\uparrow$ && Mean $\downarrow$ & Median $\downarrow$ & Err($\leq 0.5$\,m) $\uparrow$  & Err($\leq 1$\,m) $\uparrow$ && Mean $\downarrow$ & Median $\downarrow$ & Err($\leq 0.5$\,m) $\uparrow$ & Err($\leq 1$\,m) $\uparrow$ \\
		\midrule
		Associative3D~\cite{qian2020_associative3d} && \textbf{0.598} && \textbf{2.438} & 1.829 & 0.066 & 0.240 && 1.500 & \textbf{0.956} & 0.294 & \textbf{0.516} \\
	 	Ours (gt bb) && 0.380 && 2.492 & \textbf{1.775} & \textbf{0.135} & \textbf{0.292} && \textbf{1.225} & 0.995 & \textbf{0.350} & 0.504\\
	 	\bottomrule
	\end{tabular}}}
\end{table*}

\begin{table}[tb]
	\scriptsize
	\centering
	\setlength{\tabcolsep}{1.3pt}
	\newcommand{\gcs}{\hspace{5pt}}	
    \caption{\textbf{Ablation study} for ROM-feature-based matching using ground truth object detections as input. Each loss functions improves the results for certain difficulty levels.}
	\label{tab:ablation}
	\begin{tabular}{lc@{\gcs}cccc@{\gcs}cccc@{\gcs}cccc@{\gcs}ccc}
		\toprule
        && \multicolumn{3}{c}{Easy} && \multicolumn{3}{c}{Hard} && \multicolumn{3}{c}{Very Hard}\\
         	\cmidrule(l){2-5} \cmidrule(l){6-9} \cmidrule(l){10-13}
	 	&& F1$\uparrow$ & Prec$\uparrow$ & Rec$\uparrow$ && F1$\uparrow$ & Prec$\uparrow$ & Rec$\uparrow$ && F1$\uparrow$ & Prec$\uparrow$ & Rec$\uparrow$ \\
		\midrule
		only ResNet34 feats && 0.387 & 0.365 & 0.411 && 0.235 & 0.201 & 0.281 && 0.190 & 0.149 & 0.260\\
		w/o viz-input && 0.396 & \bf 0.620 & 0.291 && 0.204 & 0.389 & 0.138 && 0.178 & 0.332 & 0.122\\
		w/o pos-input && 0.437 & 0.545 & 0.365 && 0.285 & 0.402 & 0.221 && 0.262 & 0.376 & 0.201\\
        w/o AGNN && 0.456 & 0.590 & 0.372 && 0.294 & 0.411 & 0.229 && 0.272 & 0.374 & 0.214\\
        only self-att. GNN && 0.517 & 0.613 & 0.446 && 0.344 & \underline{0.452} & 0.277 && 0.292 & 0.399 & 0.231%
		\\[0.3pt] \hdashline \noalign{\vskip 2pt}
		w/o $L_{\mathit{cls}}$  && 0.525 & \underline{0.619} & 0.456 && 0.338 & \bf 0.456 & 0.268 && 0.291 & \bf 0.433 & 0.219\\
		w/o $L_{\mathit{pos}}$  && \underline{0.533} & 0.589 & \underline{0.487} && \underline{0.357} & 0.428 & \bf 0.306 && \bf 0.315 & 0.379 & \underline{0.269}\\
		w/o $L_{\mathit{dist}}$ && 0.471 & 0.478 & 0.463 && 0.322 & 0.348 & 0.299 && 0.272 & 0.242 & \bf 0.310\\
		w/o $L_{\mathit{cls/pos/dist}}$   && 0.486 & 0.574 & 0.421 && 0.330 & 0.433 & 0.267 && \underline{0.310} & \underline{0.403} & 0.252\\
		w/o $L_{\mathit{aff}}$  && 0.378 & 0.386 & 0.370 && 0.202 & 0.199 & 0.205 && 0.164 & 0.148 & 0.184
		\\[0.3pt] \hdashline \noalign{\vskip 2pt}
	 	ROM features (gt bb) && \bf 0.538 & 0.594 & \bf 0.492  && \bf 0.358 & 0.431 & \bf 0.306 && 0.308 & 0.381 & 0.259\\
	 	\bottomrule
	\end{tabular}
\end{table}
\begin{table}[tb]
	\scriptsize
	\centering
	\setlength{\tabcolsep}{6pt}
	\newcommand{\gcs}{\hspace{20pt}}
	\caption{\textbf{Object matching results} for detections predicted by EfficientDet-D7~\cite{tensorflowmodelgarden2020,huang2017_objectdetectors}. 
	We evaluate the matching for recovering all GT matchings and the matchings of the detected objects only. 
	The scores are computed for the combination of all test subsets.
	For reference, we further show results of when using GT bounding boxes.
	\label{tab:detections}}
	\begin{tabular}{lc@{\gcs}cccc@{\gcs}cc}
		\toprule
		&& \multicolumn{3}{c}{all GT matchings} && \multicolumn{2}{c}{detections only}\\
		\cmidrule(l){2-5} \cmidrule(l){6-8} 
		&& F1$\uparrow$ & Prec$\uparrow$ & Rec$\uparrow$ && F1$\uparrow$ & Rec$\uparrow$ \\
		\midrule
 		Ours (gt bb) && 0.526 & 0.552 & 0.502 && - & - \\
 		\hdashline \noalign{\vskip 2pt}
 		SuperGlue~\cite{sarlin20_superglue} && 0.139 &\bf 0.413 & 0.084 && \bf 0.385 & 0.360 \\
 		ROM features && 0.095 & 0.168 & 0.066 && 0.212 & 0.285\\
 		Ours && \bf 0.143 & 0.230 & \bf 0.103 &&  0.303 &\bf 0.445\\
	 	\bottomrule
	\end{tabular}
\end{table}

\section{Experiments}
We evaluate our method on realistically rendered indoor scenes with ground-truth object matchings.
Our experiments address three key questions:
\textbf{(1)} What is the matching performance of our approach for objects for various difficulty levels wrt. view point change?
\textbf{(2)} How does our approach relate to state-of-the-art baselines based on object and keypoint matching?
\textbf{(3)} What are the benefits of our design choices of auxiliary losses and differentiable matching?


\boldparagraph{Datasets.}
We evaluate our approach on the photorealistic synthetic Hypersim dataset of indoor scenes~\cite{roberts21_hypersim}.
We filter scenes with unrealistic appearance as well as scenes or images containing less than three objects.
We discard structural objects (e.g. wall, floor) as well as instances of the "otherprop" class and objects with minimum bounding box side length of 25 pixels.
Structural objects are typically not well localized in position, whereas the objects in the "otherprop" class are typically small in the image with only little context within the bounding box for object feature extraction.
We consider pairs of images from the same scene where at least two objects are depicted in both of them.
For our experiments, we use the official train/val/test split which results in a final number of 302/43/40 rooms and a total number of approximately 45.8k/6.3k/6.3k images.
Per training epoch, we iterate over all frames and sample a valid counterpart.
For testing, we generated three subsets of image pairs with different level of difficulty by sampling one pair per frame in the test split in each category (if available) 
We consider (a) the average difference of distances $\overline{d}$ of objects from the respective cameras as well as (b) the average over angles $\overline{\alpha}$ between the object-to-camera rays.
We denote the subsets as easy ($\overline{d}\leq 4\,m, ~\overline{\alpha}\leq 45^{\circ}$), hard (remaining objects with $\overline{d} \leq 8\,m, \overline{\alpha}\leq 90^{\circ}$), and very hard ($\overline{d} > 8\,m$ or $~\overline{\alpha}\ > 90^{\circ}$).
The subsets consist of 45.7k / 6.2k / 6.3k (easy), 42.7k / 5.8k / 6.1k (hard), and 24.4k / 3.6k / 3.0k (very hard) image pairs for training / validation / testing, respectively.
For the ablation and comparison to baselines, we provide results with GT object bounding boxes as input as a base result for an optimal detector.
We also show results on predicted object detections from EfficientDet-D7~\cite{tensorflowmodelgarden2020,huang2017_objectdetectors} trained on the MS COCO dataset~\cite{lin2014_mscoco}.

\boldparagraph{Network Parameters and Training Details.}
We apply a pretrained ResNet34~\cite{he2016_resnet} to obtain object-wise visual features from 128$\times$128 patches obtained by rescaling the detected object bounding boxes.
Tab.~\ref{tab:params} lists the network parameters of object encoder and matching network.
The AGNN alternates self- and cross-attention 2 times. 
During training, we add Gaussian noise with zero mean and variance 0.01 to the precomputed visual feature vectors for data augmentation.
We only consider a maximum number of 40 objects per frame and sample objects if more are present.
Input bounding box coordinates are normalized wrt. image size to values between 0 and 1.
The loss weights are set to $\{\lambda_{\mathit{aff}}, \lambda_{\mathit{cls}}, \lambda_{\mathit{pos}}, \lambda_{\mathit{rel}}\} = \{1, 1, 0.1,0.1\}$.
We found 10 Sinkhorn iterations to be sufficient to converge.
We train our model using ADAM~\cite{kingma2014_adam} with learning rate $10^{-4}$ and batch size 32 for 350 epochs.

\begin{figure*}[th!]
	\scriptsize
	\centering
	\setlength{\tabcolsep}{1pt}
	\renewcommand{\arraystretch}{0.6}
	\newcommand{\sz}{0.32}
	\begin{tabular}{cccc}
		& Easy & Hard & Very Hard
		\\
		\rotatebox{90}{\hspace{7pt}SuperGlue~\cite{sarlin20_superglue}} &
        \includegraphics[width=\sz\textwidth]{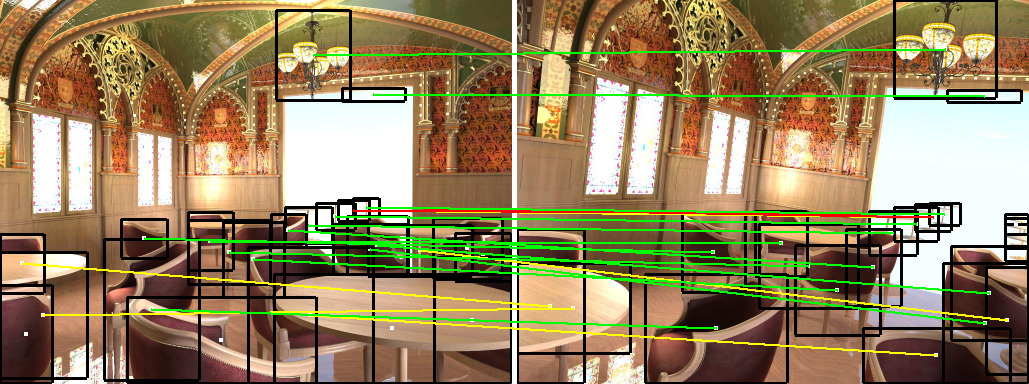}&
		\includegraphics[width=\sz\textwidth]{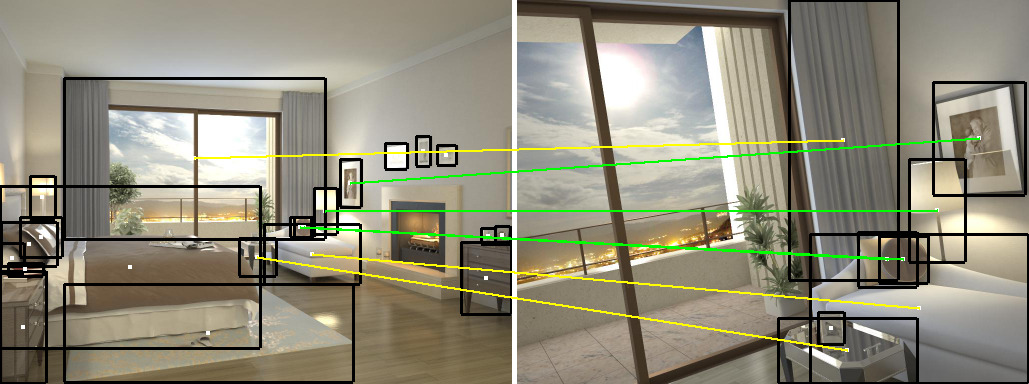} &
		\includegraphics[width=\sz\textwidth]{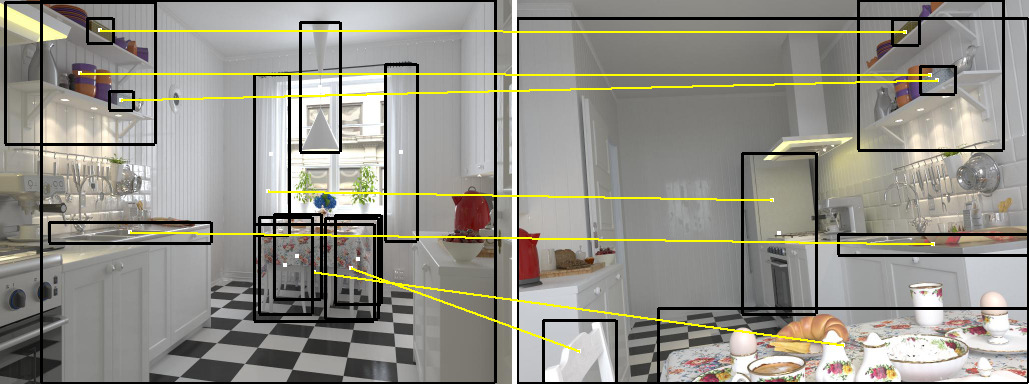} 
		\\[1pt]
        \rotatebox{90}{\hspace{8pt}ROM features} &
		\includegraphics[width=\sz\textwidth]{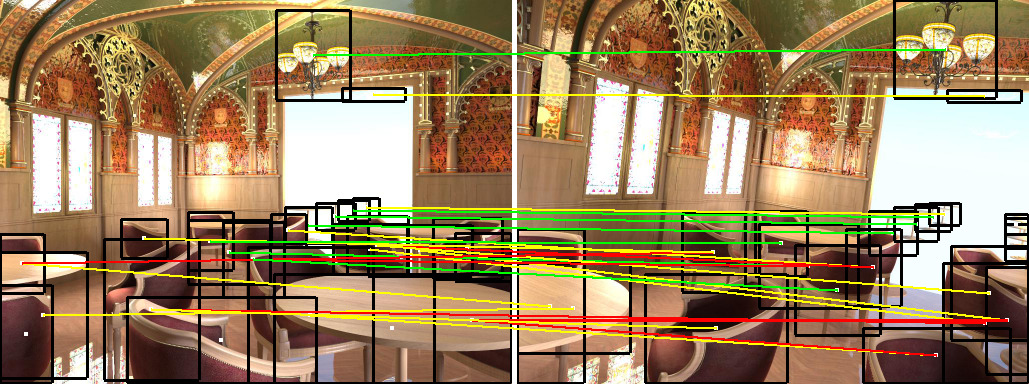}&
		\includegraphics[width=\sz\textwidth]{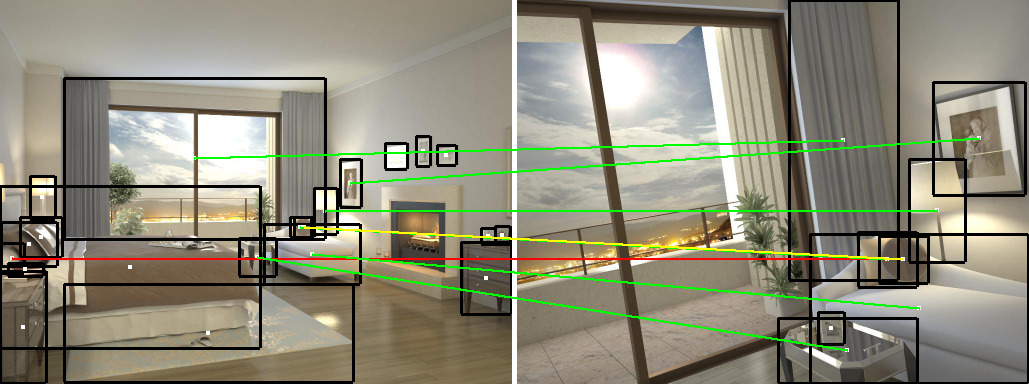} &
		\includegraphics[width=\sz\textwidth]{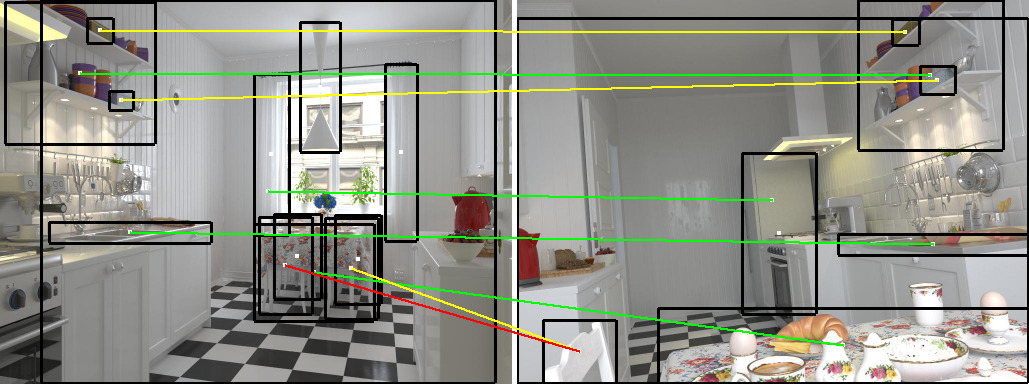} 
		\\[1pt]
		\rotatebox{90}{\hspace{22pt}Ours} &
		\includegraphics[width=\sz\textwidth]{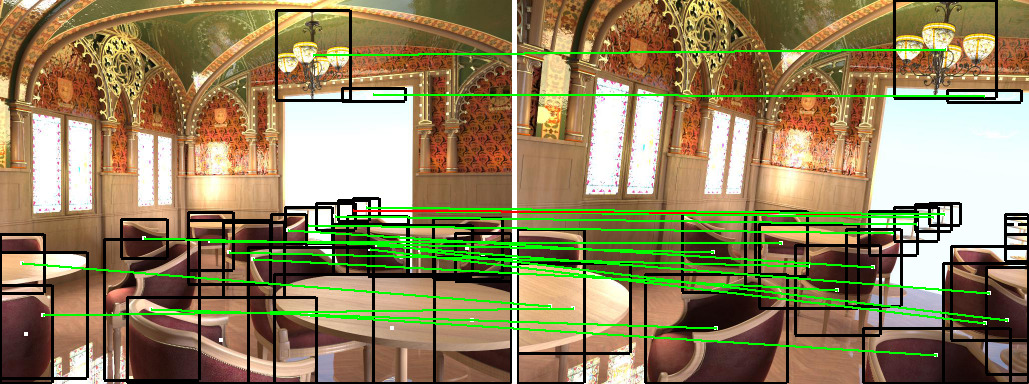} &
		\includegraphics[width=\sz\textwidth]{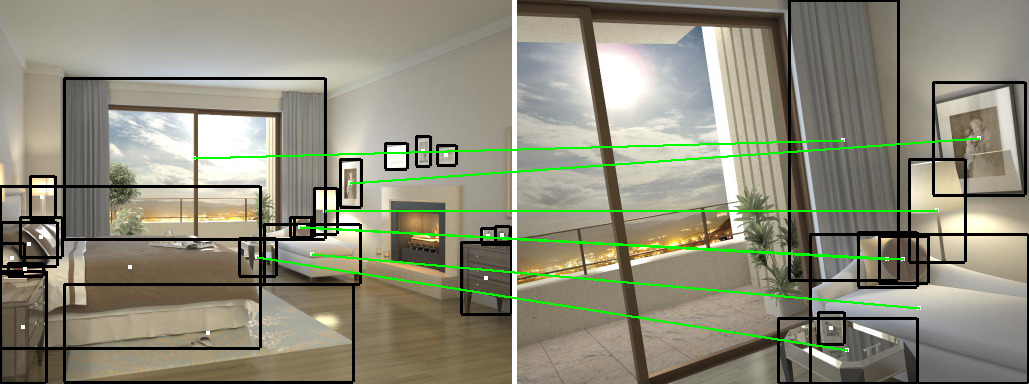} &
		\includegraphics[width=\sz\textwidth]{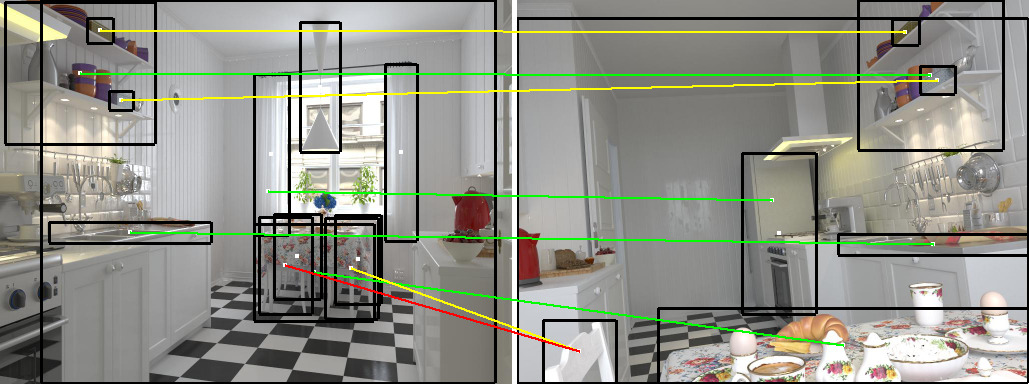} 
        \\[-4pt]
	\end{tabular}
	\caption{\textbf{Qualitative comparison.} Evaluation on GT bounding boxes; \textbf{green:} correctly detected match, \textbf{red:} wrongly detected match, \textbf{yellow:} missing match.
	While SuperGlue yields very accurate matching results for the examples of easy and hard view point changes, it cannot detect any matches if the change of view point is too large (right column, very hard).
	On the other hand, our ROM-features can also be used to match objects in images with larger changes.
	Combining both object- and keypoint-based matching benefits from both capabilities.
	}
	\label{fig:baselines}	
\end{figure*}

\begin{figure*}[th!]
	\scriptsize
	\centering
	\setlength{\tabcolsep}{1pt}
	\renewcommand{\arraystretch}{0.6}
	\newcommand{\sz}{0.32}
	\newcommand{\gcs}{\hspace{5pt}}	
	\begin{tabular}{ccc@{\gcs}c}
		\includegraphics[width=\sz\textwidth]{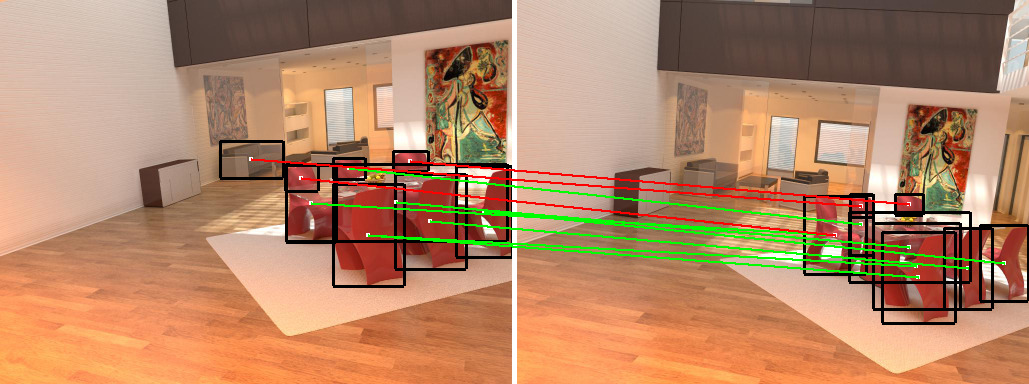}&
		\includegraphics[width=\sz\textwidth]{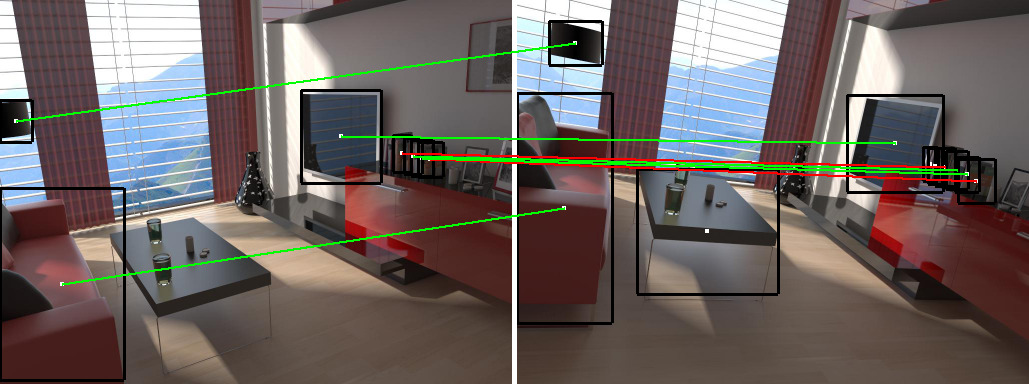} &&
		\includegraphics[width=\sz\textwidth]{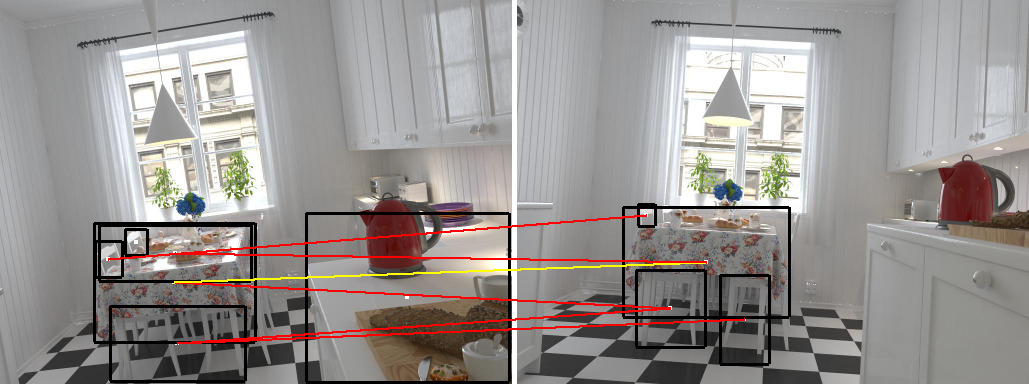} 
	\end{tabular}
	\caption{\textbf{Qualitative results from our method.} Evaluation on detected bounding boxes; \textbf{green:} correctly detected match, \textbf{red:} wrongly detected match, \textbf{yellow:} missing match. Right: failure case with inaccurately detected object bounding boxes.}
	\label{fig:detections}	
\end{figure*}


\boldparagraph{Evaluation Metrics.}
The main task of our approach is to obtain valid object matches in pairs of images.
Given the assignment matrix $\overline{\mathbf{P}}$, we determine a fixed assignment per object for both images either to some object of the other image or to the dustbin by maximum value assignment.
We compute the recall of correct object-object matches over all ground truth matches.
Precision is defined as the ratio of correct object-object matches among all found object-object matches.
We either compute the respective measure over all objects in all images combined (\textit{object-wise}) or individually over all objects per frame-pair with subsequent averaging over all images (\textit{frame-wise}). 
Please note that scenes with a very large number of objects have a high impact on the first-named variant.
When evaluating our approach using objects from a detector network as input, the detector might detect different objects than actually available in the ground truth. 
Hence, we only evaluate the matching for detections which are assigned to a ground truth object in the same image. 
A detection is assigned to a ground truth object if it overlaps and has maximal intersection-over-union with the ground truth bounding box.
We further evaluate the performance of our models on the auxiliary tasks.
For the classification task, we compute the accuracy of the semantic predictions over all object.
Similar to \cite{qian2020_associative3d}, we compute mean and median errors, and the rate of estimates with errors $\leq 0.5/1.0\,m$ for both the distance and position prediction.


\boldparagraph{Baselines.}
We compare our method with two state-of-the-art baselines, CSR~\cite{gadre2022_csr} and Associative3D~\cite{qian2020_associative3d}, which learn object matching from bounding boxes.
We trained the approaches on our dataset using the public reference implementation until convergence on the validation set.
We further use SuperGlue~\cite{sarlin20_superglue} keypoints to determine an object-level matching by applying Sinkhorn iterations on $\mathbf{S}^{\mathit{kp}}$.


\subsection{Object Matching Results}

\boldparagraph{Ground-truth detections.}
In \reftab{gtmatching} we show matching results of our approach using ground truth object bounding boxes as input. 
We also compare our method with state-of-the-art approaches which are trained end-to-end for object description and matching (CSR, Associative3D) and a keypoint-based baseline (SuperGlue). 
It can be observed that both using only our ROM features as well as our combined method consistently outperforms CSR and Associative3D.
SuperGlue has an advantage on smaller to medium (easy) view point changes, while our ROM features are on par (hard) or better (very hard) on larger view point changes in recall and F1 score.
Especially, we found it challenging for SuperGlue to detect any keypoint matches in case of larger view point changes, e.g. when the same scene was depicted from opposite sides.
By combining the two matching principles, we can benefit from the advantages of both methods.
In most cases, inference and matching time for our ROM features took approx. $50-100$ ms per image pair.
We show qualitative examples and comparisons with baselines in Fig.~\ref{fig:baselines}.
In \reftab{sidetasks}, we further evaluate the performance of our network on the auxiliary tasks of classification, 3D position, and distance estimation and compare with Associative3D.
Our approach estimates 3D position and distance of objects with a median error of 1.775\,m and 0.995\,m, respectively, from RGB images.


\reftab{ablation} shows results of various ablations of our model using ground truth object detections as input.
Each architectural design choice contributes to the overall performance.
The classification and 3D losses do only bring little improvements on the easy and hard cases, while the distance loss helps to improve the matching performance on all difficulty levels.
Also both self- and cross-attention and using visual (viz-input) or position (pos-input) features are important for the performance of the method.


\boldparagraph{Predicted object detections.}
We also demonstrate our approach for predicted object detections in \reftab{detections}.
From the detections, we remove those whose predicted class label (COCO labels) does not appear in the NYU40~\cite{silberman2012_nyu} class labels used in our GT. 
We further only consider detections with a minimum detection score of 0.3.
Predicted bounding boxes are assigned to an associated GT object if the intersection over union (IoU) between GT and prediction is greater than 0.5.
If multiple objects would get assigned to the same objects, we choose the one with highest IoU.
By this, approx. $40\%$ of the detections are assigned to GT objects and $30\%$ of all GT objects are found.
Contrary to using GT bounding boxes, less than three objects might get found in an image and forwarded to the model.
We report F1 and recall scores for recovering the matching for the detected objects that have GT correspondence.
We also show results for recovering the complete GT matching which, however, depends on the detector performance too.
For matching available predicted detections, our approach achieves moderately lower scores than for matching GT detections.

\subsection{Limitations and Future Work}
Our matching approach relies on a pretrained detector and the quality of the detected bounding boxes.
In future work, end-to-end training of the detector with the matching pipeline could be investigated.
In difficult occlusion settings, object bounding boxes can overlap and similar features are computed. 
Calculating object features based on instance segmentation could be interesting for future research.
In this work, we focused on photorealistic synthetic, static indoor scenes with full 3D object-wise position information during training which allows our full model to consider distances between objects.
While our model also achieves decent performance without considering spatial information of objects, we believe that ideas for matching objects in case of scene changes (e.g. outdoor scenes with moving cars), different camera geometries, or limited supervision are further interesting directions to explore.

\section{Conclusion}

We proposed a novel approach for matching 2D object detections between images.
Our approach is trained to match objects across a large variety of view points with auxiliary classification and 3D regression tasks.
This way, our approach can outperform keypoint-based matching approaches for large view point changes.
We evaluate our approach using a realistically rendered indoor dataset, and also demonstrate state-of-the-art performance among approaches which train object-wise matching end-to-end.
Future applications of our approach could be explored for tasks such as object-centric scene reconstruction, image retrieval, and localization.

\addtolength{\textheight}{-3cm}   


%


\bibliographystyle{IEEEtranS}
\bibliography{egbib}

\end{document}